\documentclass[sigplan,screen]{acmart}
\usepackage{graphicx}
\usepackage{booktabs}
\usepackage{textcomp}
\usepackage{xcolor}
\usepackage{tikz}
\usepackage{enumitem}
\usepackage{multirow}
\usepackage{amsmath}
\usepackage{graphicx}
\usepackage{subfigure}
\usepackage{makecell}
\usepackage{amsmath}
\usepackage{mathtools}
\usepackage{amsthm}
\usepackage{hyperref}

\usepackage{soul}

\usepackage{xcolor}
\usepackage{colortbl}
\usepackage{booktabs}

\usepackage{algpseudocode}
\usepackage{algorithm}

\usepackage{setspace}

\AtBeginDocument{%
  \providecommand\BibTeX{{%
    \normalfont B\kern-0.5em{\scshape i\kern-0.25em b}\kern-0.8em\TeX}}}

\usepackage{dblfloatfix}
\author[1]{{\normalsize Hanqiu Chen$^1$, Xuebin Yao$^2$, Pradeep Subedi$^2$ and Cong Hao$^1$}
\\
{\normalsize$^1$Georgia Institute of Technology and $^2$Samsung Semiconductor, Inc.}\\
{\normalsize\{hanqiu.chen, callie.hao\}@gatech.edu, \{xuebin.yao, prad.subedi\}@samsung.com}
}
\copyrightyear{2024}
\acmYear{2024}
\setcopyright{rightsretained}
\acmConference[ICCAD '24]{IEEE/ACM International Conference on Computer-Aided Design}{October 27--31, 2024}{New York, NY, USA}
\acmBooktitle{IEEE/ACM International Conference on Computer-Aided Design (ICCAD '24), October 27--31, 2024, New York, NY, USA}
\acmDOI{10.1145/3676536.3676673}
\acmISBN{979-8-4007-1077-3/24/10}

\renewcommand{\shortauthors}{Chen et al.}
\begin{document}

\title{  Residual-INR: Communication Efficient On-Device Learning Using Implicit Neural Representation}

\begin{abstract}



Edge computing is a distributed computing paradigm that collects and processes data at or near the source of data generation. The on-device learning at edge relies on device-to-device wireless communication to facilitate real-time data sharing and collaborative decision-making among multiple devices. This significantly improves the adaptability of the edge computing system to the changing environments. However, as the scale of the edge computing system is getting larger, communication among devices is becoming the bottleneck because of the limited bandwidth of wireless communication leads to large data transfer latency. 
To reduce the amount of device-to-device data transmission and accelerate on-device learning, 
in this paper, we propose Residual-INR, 
a fog computing-based communication-efficient on-device learning framework by utilizing implicit neural representation (INR) to compress images/videos into neural network weights. 
Residual-INR enhances data transfer efficiency by collecting JPEG images from edge devices, compressing them into INR format at the fog node, and redistributing them for on-device learning. By using a smaller INR for full image encoding and a separate object INR for high-quality object region reconstruction through residual encoding, our technique can reduce the encoding redundancy while maintaining the object quality.
Residual-INR is a promising solution for edge on-device learning because it reduces data transmission by up to 5.16 $\times$ across a network of 10 edge devices. It also facilitates CPU-free accelerated on-device learning, achieving up to 2.9 $\times$ speedup without sacrificing accuracy. Our code is available at: \url{https://github.com/sharc-lab/Residual-INR}.

\end{abstract}

\maketitle

\section{Introduction}
\label{sec:intro}

\begin{figure}[t!]
\centering
\subfigure{\includegraphics[width=1.00\linewidth]{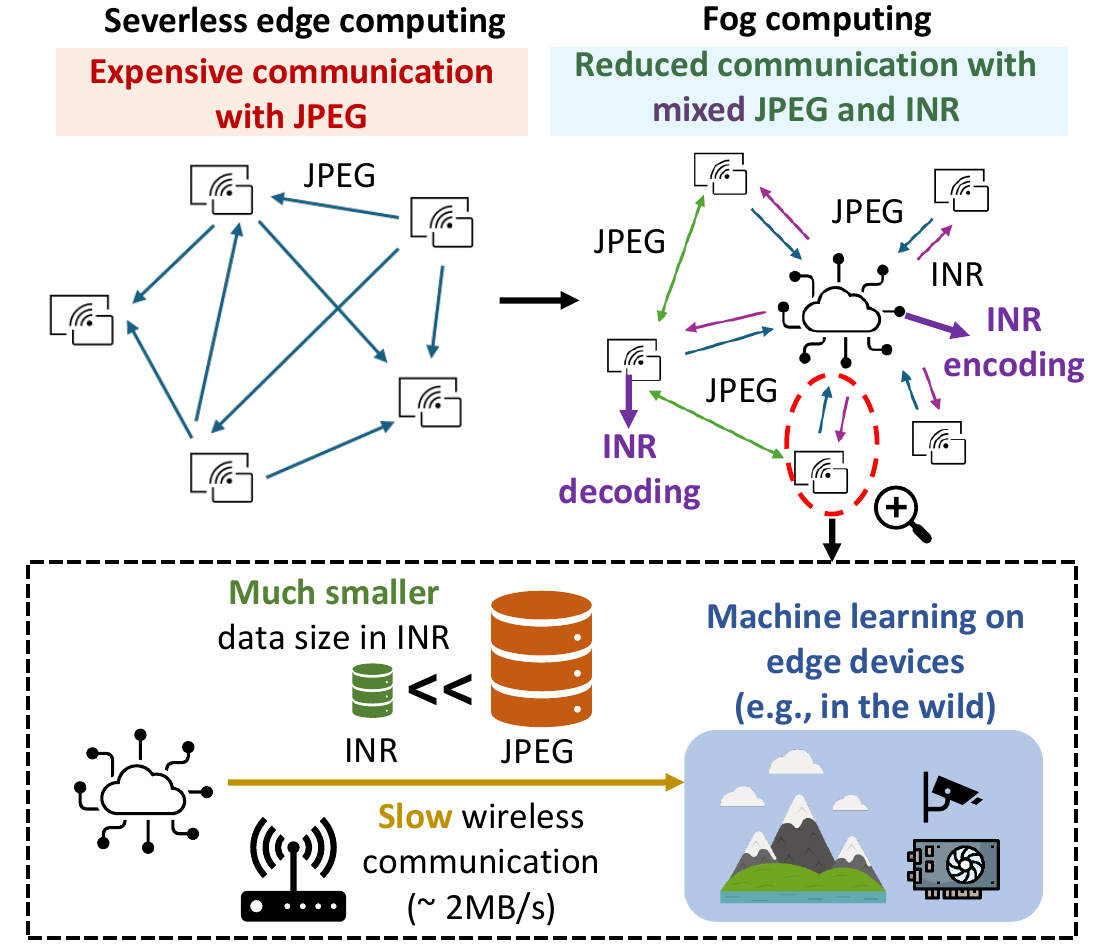}}
\vspace{-18pt}
\caption{Using implicit neural representation for data compression in fog on-device learning reduces amount of wireless data transmission.}
\label{fig:1_background_overview}
\vspace{-15pt}
\end{figure}

AI workloads are increasingly shifting from centralized cloud architectures to distributed edge systems to process data closer to where it is generated~\cite{hao2021enabling,shi2016edge, zhou2019edge, wang2020convergence, xu2020edge}. On-device learning, by leveraging these edge systems, offers enhanced real-time processing capabilities with adaptation to the changing environments~\cite{sunaga2023addressing, zhu2022device}. Severless edge computing, as shown in Fig.~\ref{fig:1_background_overview}, is one of the most commonly used computing paradigm. As edge devices continually collect new data, AI workloads on these devices need to be updated to adapt to emerging new tasks. However, this approach necessitates data sharing and synchronization across devices to ensure that all edge devices have access to the most up-to-date data for local computation. Nonetheless, large-scale serverless edge computing can result in substantial device-to-device communication burdens. As shown in Fig.~\ref{fig:1_background_overview}, for edge devices that are deployed in the wild, those device-to-device communication relies on wireless communication, which has a very limited bandwidth. For example, the bandwidth provided by 4G LTE~\cite{Verizon2021} is only about 5-12 Mbps, causing a large latency with heavy communication. 

To optimize data management and communication among edge devices, fog computing has been proposed~\cite{bonomi2012fog, sarkar2019serverless}. This technique integrates a fog node with 
higher computational and storage capacities than edge devices into the network, allowing for the offloading of intensive computation tasks and data compression~\cite{sadri2022data, naeem2019fog, sarkar2015assessment}. However, the substantial communication traffic between the fog node and edge devices remains a critical concern, especially for data-intensive computer vision tasks. Image compression offers a viable solution to reduce heavy communication loads. However, recent classical or neural image compression algorithms are computation-intensive~\cite{dubois2021lossy, singh2020end, johnston2018improved, theis2022lossy}, or optimized for perceptual quality~\cite{balle2016end, minnen2018joint}, making them unsuitable for resource-constrained edge devices and causing redundant compression for machine learning training.


Recently, implicit neural representation (INR)~\cite{sitzmann2020implicit, dupont2021coin, dupont2022coin++, chen2023rapid, chen2021nerv} is emerging as a novel image compression technique by training a neural network to effectively encoding the entire image as a compact set of network parameters that can be easily stored and reconstructed. To use INR compressed images for downstream machine learning tasks, Rapid-INR~\cite{chen2023rapid} demonstrates its success on image classification with higher compression rate than JPEG. However, whether INR compression can be extended for  communication-efficient edge on-device computer vision tasks
has not been explored yet. As shown in Fig.~\ref{fig:1_background_overview}, since INR compressed images can be much smaller than JPEG, compressing images into INR format at the fog node for data transmission among devices can save a large amount of data transmission.

Motivated by the need for reducing communication among edge devices and the unexplored potential of INR in communication efficient edge computing, we propose \textbf{Residual-INR}, a fog computing-based communication-efficient on-device learning framework by utilizing INR to compress images/videos. 
Utilizing the fog node for encoding background and object to INR separately in different quality, Residual-INR remarkably reduces data transmission among edge devices while maintaining object encoding quality, thereby accelerating on-device training without large accuracy loss. Our contributions can be summarized as follows:

\begin{enumerate}[leftmargin=*]

\item \textbf{System - Efficient fog online learning with hybrid JPEG-INR communication.} To improve communication efficiency in distributed edge online learning, we propose using fog computing 
instead of serverless edge computing with pure JPEG image transmission. In our proposed fog computing system, the fog node collect images in JPEG from edge devices, compress them into INR format, and redistribute them for on-device learning. INR images are decoded on-the-fly on edge devices during training. Our hybrid JPEG-INR transmission strategy considerably reduces data communication and accelerates transmission.


\item \textbf{Algorithm - Versatile region importance aware INR compression for reduced encoding redundancy.} We propose region importance aware INR encoding, which contrasts with traditional single INR encoding that assigns equal importance to all pixels. We utilize a smaller \emph{background INR} developed upon previous INR networks for compressing the entire image at a lower quality. A separate \emph{object INR} enhances the encoding quality of objects through residual encoding. 
This approach not only reduces the combined size of the background INR and object INR compared to a single INR but also preserves object detection training accuracy. Additionally, our technique is versatile and can be integrated seamlessly with existing INR compression networks.


\item \textbf{Hardware - CPU-free INR decoding with workload balancing.} To address workload imbalances from varying decoding latencies due to different INR sizes, we group images with the same sized INR together for parallel decoding during training on edge device. Additionally, the compact size of INR weights allows them to be stored within device memory, enabling CPU-free training without frequent external storage access and eliminating the need for a complex software stack.



\item \textbf{ Mathematical modeling - An analytical math model for optimal communication strategy.} 
We develop a mathematical model to model the data communication in the whole system.
 We identify the optimal compression and communication strategy—whether to send JPEG images to the fog node for INR compression or directly share JPEG images with other devices. Additionally, we determine the most communication efficient locations for training—whether at the fog node or the edge.



\item \textbf{Evaluation - Thorough experiment analysis.} 
We conduct a thorough experiment to demonstrate the advantages of Residual-INR. 
Compared with JPEG compression, Residual-INR reduces the average image size by up to 12.1 $\times$ with similar object encoding quality.
In an edge computing network with 10 devices, Residual-INR reduces data transmission amount
by up to 5.16 $\times$ and accelerates end-to-end training time by up to 2.9 $\times$, without sacrificing accuracy.

\end{enumerate}

\section{Preliminary and Motivations}

\subsection{Implicit neural representation for image and video compression}
\label{Sec: Rapid-INR and NeRV}


INR
uses neural networks to model complex hidden features, presenting a novel method for parameterizing various data types~\cite{sitzmann2020implicit, dupont2021coin, dupont2022coin++, chen2023rapid, chen2021nerv}. The application of INR in data compression involves training a neural network to converge and then compressing the weights to reduce its size. This technique has gained increased attention in computer vision for compressing images and videos. For instance, COIN~\cite{dupont2021coin}, COIN++~\cite{dupont2022coin++} and Rapid-INR~\cite{chen2023rapid} employ a  multilayer perceptron (MLP) to compress images, encoding the relationship between pixel locations (x, y) and their RGB values, with each image having a dedicated MLP. For video compression, by utilizing similarities between adjacent video frames for a higher compression rate, NeRV~\cite{chen2021nerv} utilizes a single network combining MLPs and CNNs to encode the entire video sequence, taking frame indices as inputs and outputting corresponding RGB values. Different frames within a video sequence share the same network, while each video sequence is encoded by a dedicated network.

\textbf{Motivated by the higher compression rate of INR compared to JPEG}, with minimal quality loss, INR is a promising compression technique for \textbf{efficient communication in fog computing}. 


\subsection{Low object reconstruction quality}
\label{Sec:Low object reconstruction quality}

 Existing INR methods~\cite{dupont2021coin, dupont2022coin++, chen2023rapid, chen2021nerv} usually encode one image using a single INR; however, for images with less important background and more important objects, we observe that important regions usually suffer from quality degradation. Larger INRs can improve image quality, like demonstrated in Rapid-INR; however larger INRs consume larger memory and sometimes is not better than JPEG.
 
\begin{figure}[t!]
\centering
\subfigure{\includegraphics[width=1.00\linewidth]{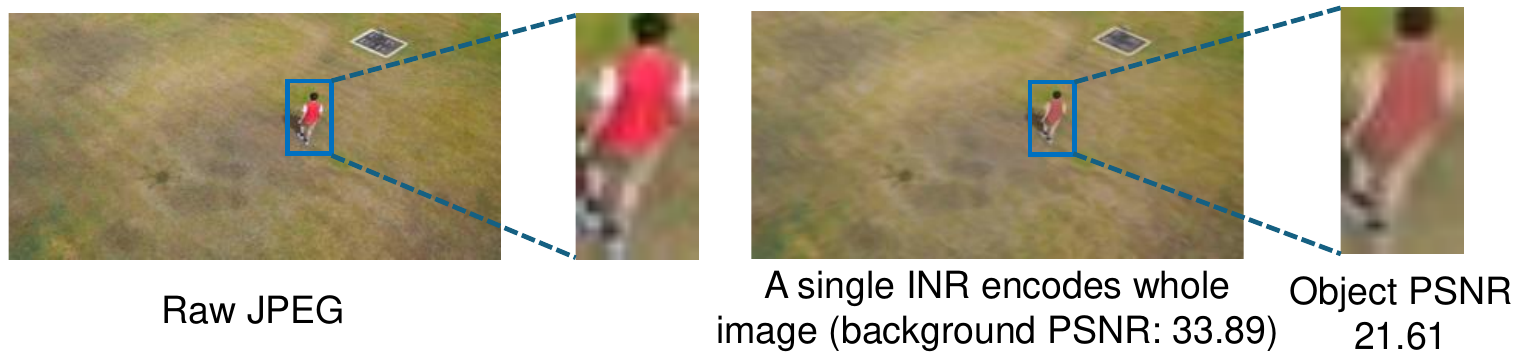}}
\vspace{-20pt}
\caption{Small object suffers from quality degradation using a single INR to encode an entire image.}
\vspace{-15pt}
\label{fig:2_motivation_visualization}
\end{figure}

\begin{figure}[t!]
\centering
\subfigure{\includegraphics[width=1.00\linewidth]{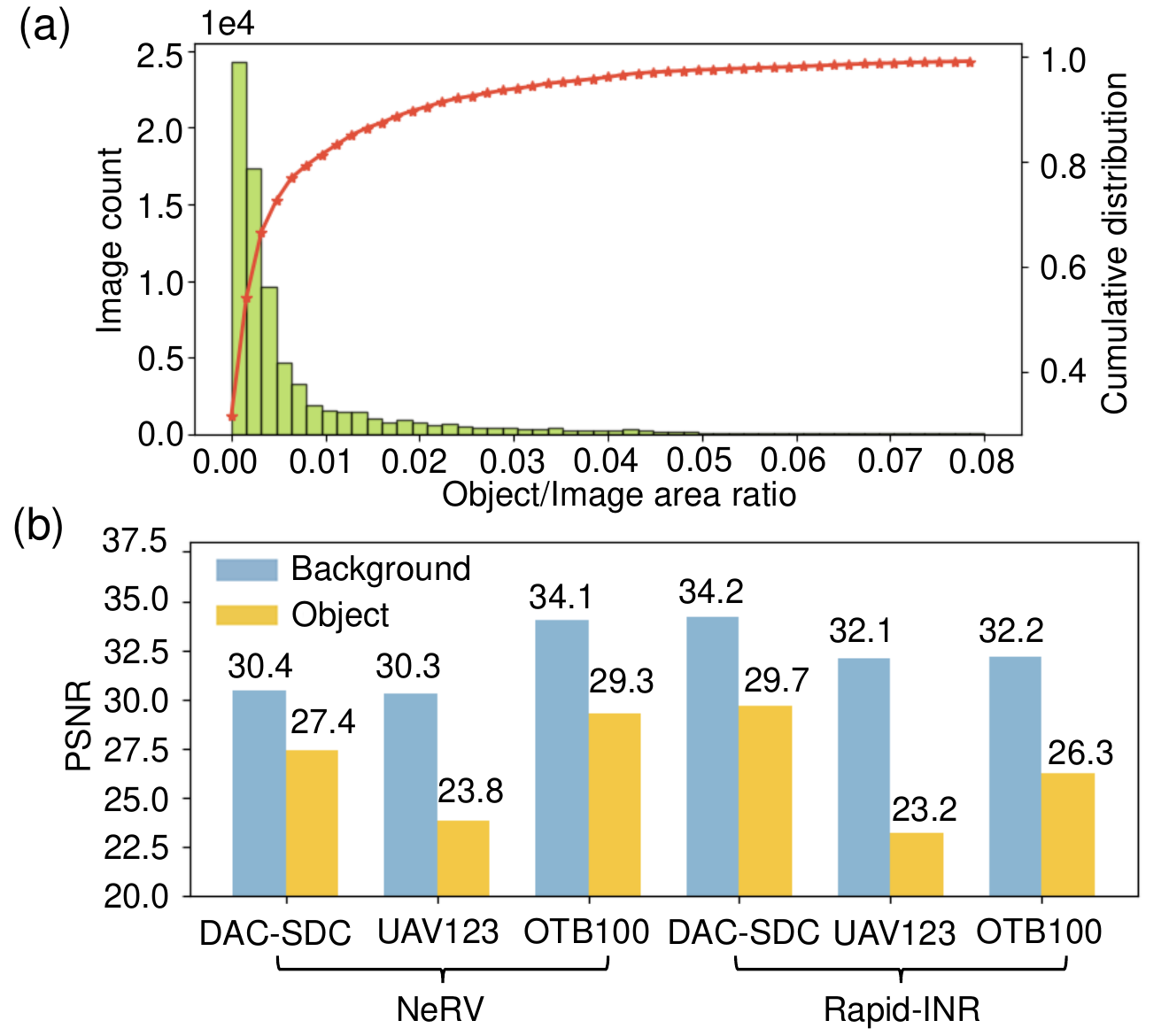}}
\vspace{-20pt}
\caption{(a) Object sizes distribution in the UAV123 dataset. (b) Comparison between the PSNR of object and background.  }

\label{fig:3_low_object_PSNR}
\vspace{-15pt}
\end{figure}

 Fig.~\ref{fig:2_motivation_visualization} shows that compressing an entire image using a single INR causes object colors to blend with the background, resulting in blurred objects that are difficult to detect. Moreover, in certain object detection tasks, objects often occupy a very small portion of the image, as presented in Fig.~\ref{fig:3_low_object_PSNR} (a). Using a single INR to encode the entire image assigns equal importance to all pixels, resulting in reduced attention to objects during encoding and lower object reconstruction quality.  We evaluate reconstruction quality using the Peak Signal-to-Noise Ratio (PSNR), where higher PSNR values indicate better quality. As a case study, Fig.~\ref{fig:3_low_object_PSNR} (b) illustrates the averaged PSNR for background and object regions across the DAC-SDC~\cite{xu2019dac}, UAV123~\cite{mueller2016benchmark}, and OTB100~\cite{wu2013online} datasets using NeRV and Rapid-INR for encoding. Notably, the PSNR for objects is significantly lower than that for backgrounds.

\textbf{Motivated by the lower reconstruction quality of objects}, which significantly affects the accuracy of detection backbone training, \textbf{we propose an additional tiny INR dedicated solely to enhancing object encoding quality}.

\section{Residual-INR Encoding and Decoding}

\begin{figure*}[t!]
\centering
\subfigure{\includegraphics[width=1.00\linewidth]{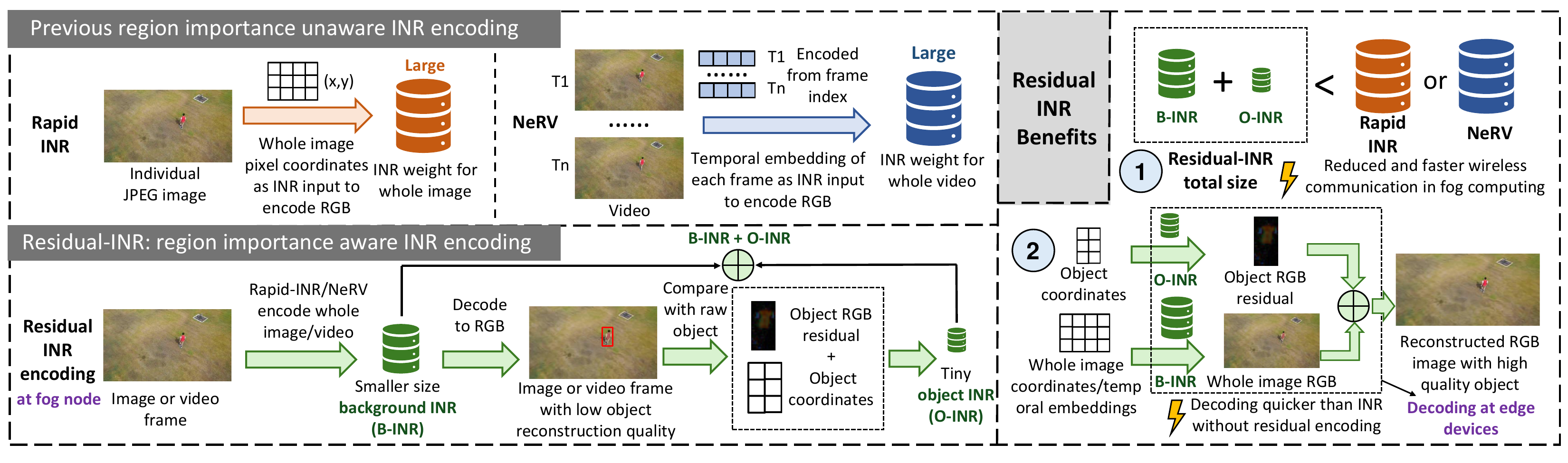}}
\vspace{-20pt}
\caption{Overview of Residual-INR encoding and decoding framework. Residual-INR consists of a background INR and an object INR. The reduced INR size facilitates faster communication and decoding speeds.}
\label{fig:4_Residual_INR_overview}
\vspace{-10pt}
\end{figure*}

Our proposed Residual-INR, similar to typical INRs, has two stages for compression: encoding an image/video into its INR format, and decoding the INR back to its original format. 
This two-stage approach is particularly suitable for fog computing, where encoding occurs at fog nodes while decoding takes place at edge devices. This is because the INR encoding process is computationally intensive due to the need for training a neural network to converge.
INR decoding is a fast and lightweight process, which even can be effectively handled by the resource-limited edge devices. As a result, transmitting data in compressed INR format from the fog node to the edge device can significantly reduce communication traffic. Residual-INR encoding builds upon existing INR encoding networks. In this paper, as a case study, we choose Rapid-INR and NeRV (Sec.~\ref{Sec: Rapid-INR and NeRV} and Fig.~\ref{fig:4_Residual_INR_overview}) as base networks for image encoding and video sequence encoding. These are referred to as \textbf{Res-Rapid-INR} and \textbf{Res-NeRV}, respectively. Note that the application of Residual-INR extends beyond these two INR networks, and can be easily adapted to other existing INR encoding frameworks with minimal modifications. Residual INR decoding is parallelized and CPU-free without the support of complex software stack, and can be accelerated by embedded GPUs or machine learning accelerators on the edge device.

\subsection{Region importance aware encoding}

Considering the fact that the background is less critical than the object in object detection training, encoding the background at a lower quality minimally impacts training accuracy while reduces encoding redundancy. Residual-INR optimizes encoding efficiency for smaller datasets size by differentiating between background and object encoding.
As shown in Fig.~\ref{fig:4_Residual_INR_overview}, it refines existing INR frameworks by utilizing a smaller INR (\textbf{background INR}) for the entire image and adding an additional tiny INR (\textbf{object INR}) specifically to enhance object region encoding quality. Background INR and object INR work together during INR decoding for image reconstruction. Background INR first decodes the entire image, followed by object INR which specifically decodes and overlays the object onto the image as a patch.
The goal is to \ul{\textit{make the size of background INR and object INR together smaller than the size of a single INR without sacrificing object reconstruction quality}}.


\begin{figure}[t!]
\centering
\subfigure{\includegraphics[width=1.00\linewidth]{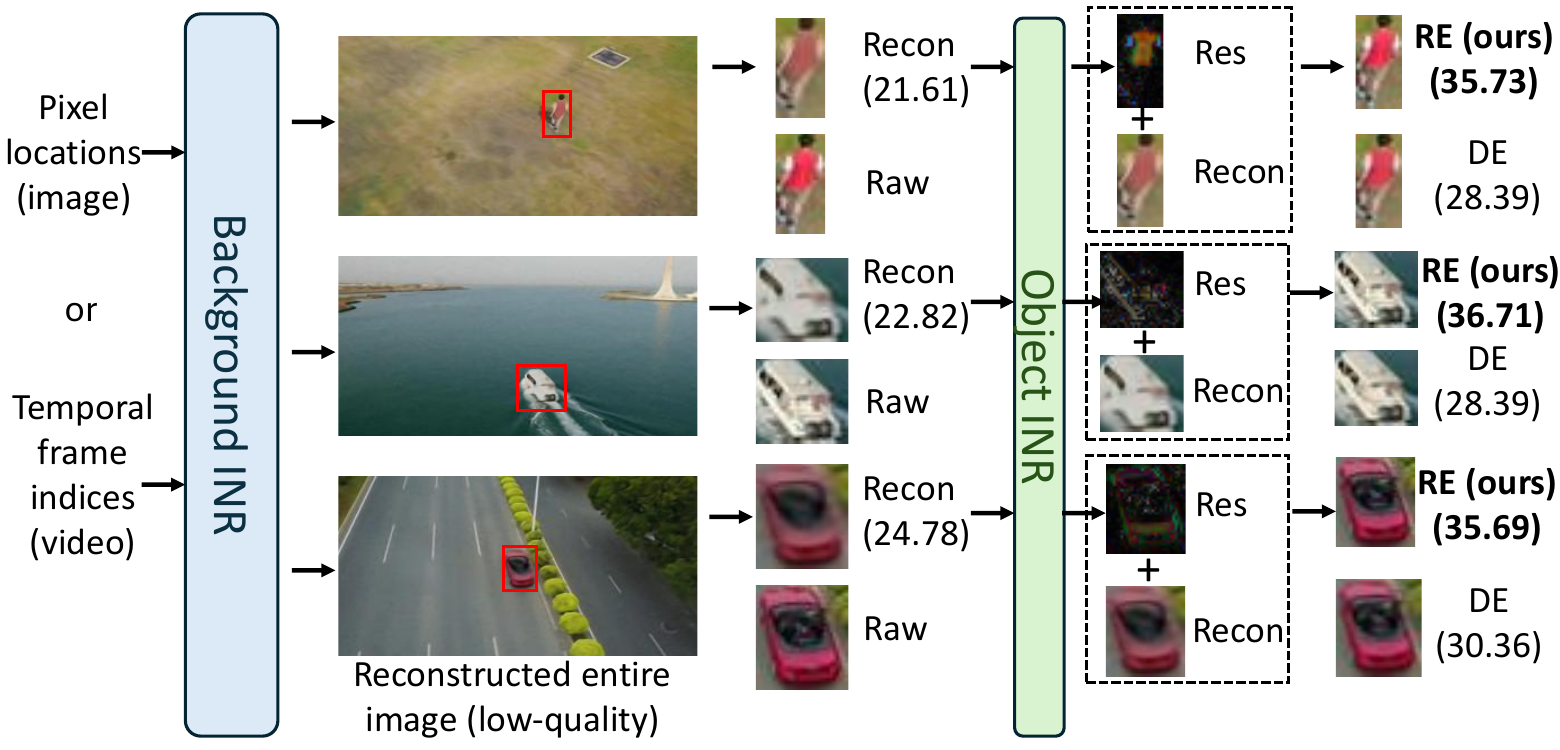}}
\vspace{-20pt}
\caption{ Visualization of Residual-INR encoding. Res: image residual. RE: residual encoding. DE: direct encoding. }
\label{fig:12_more_visualization}
\vspace{-15pt}
\end{figure}

\subsubsection{Background INR} 
Our background INR, developed upon Rapid-INR for images and NeRV for videos, uses a reduced size network to encode the entire image/video frame at a relatively lower quality that will not affect the object detection training accuracy. As a result, the smaller size background INR effectively minimizes redundancy in background encoding. Besides the network structure, all other configurations remain unchanged. Given the uniformity of image sizes within a dataset, the same sized Rapid-INR is used as background INR encoding. However, video sequences often vary in length in the dataset. To balance the compression rate and encoding quality, we employ differently sized NeRV as encoders according to the length of each video sequence. Although objects are decoded in low quality by background INR, the object features learned by backround INR will be utilized by obejct INR.

\subsubsection{Object INR} 
The object INR is used to enhance the encoding quality of the object region in the format of MLPs. The object region is identified by the bounding box boundaries. Our preliminary analysis shows that the object size is typically much smaller than the overall image size (Sec.~\ref{Sec:Low object reconstruction quality}), allowing us to use a tiny object INR for high-quality, storage-efficient encoding. Objects usually vary in size in different images. To ensure each object is encoded optimally, we use various sizes of object INRs matched to the size of each object. 

There are two ways of encoding the objects: \emph{direct encoding} using raw RGB and \emph{residual encoding} using residual RGB values. Direct encoding is to simply take the object pixel coordinates as inputs and outputs the RGB values of object pixels. The output directly replaces the low-quality object region in the background INR reconstructed image. However, this method cannot fully utilize the information that the background INR has already learned, also causing some encoding redundancy. To better leverage the already learned information by background INR, we use residual encoding. As shown in Fig.~\ref{fig:4_Residual_INR_overview}, residual encoding starts by cropping the object region RGB values decoded by background INR and its pixel coordinates from the full image. Next, we calculate the residual RGB values by comparing the reconstructed object from the background INR to the object in the raw image. The INR learning objective then shifts from matching the raw RGB values to fitting these residual values.

\begin{figure}[t!]
\centering
\subfigure{\includegraphics[width=1.00\linewidth]{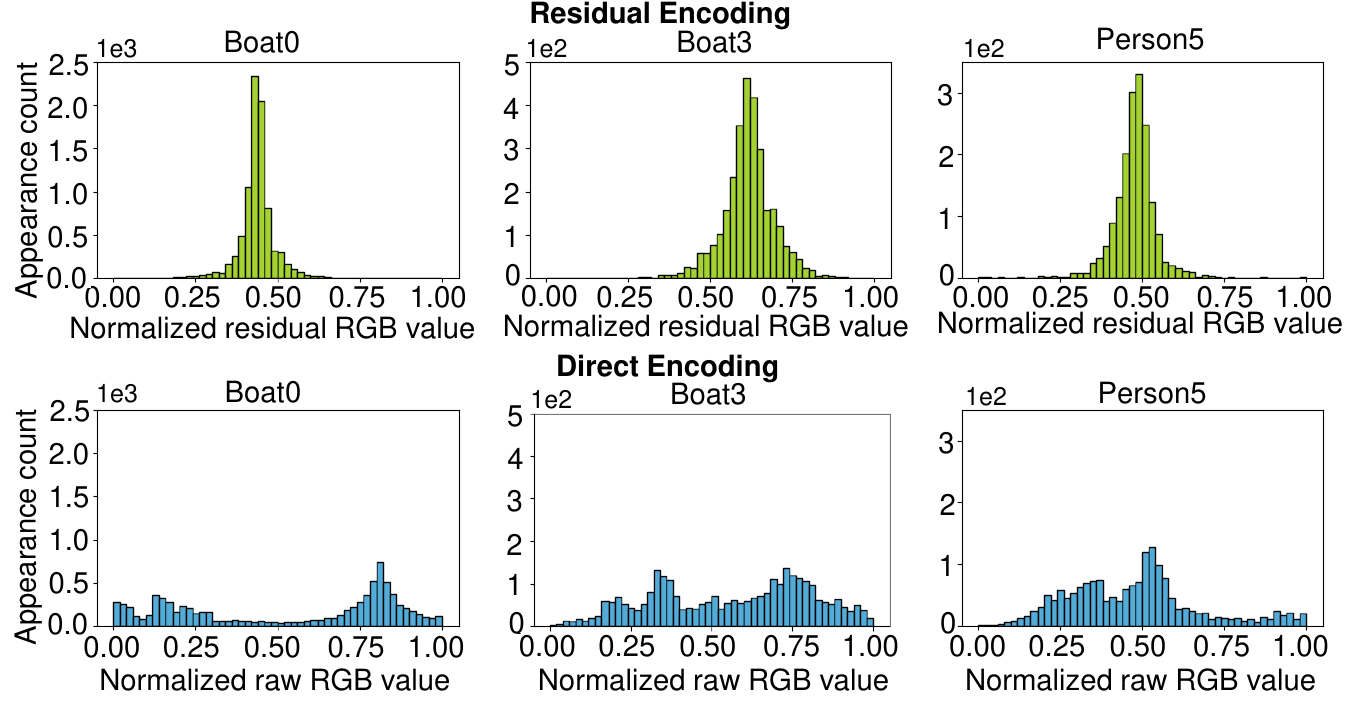}}
\vspace{-20pt}
\caption{ Comparison between normalized residual RGB values and raw RGB values. }
\label{fig:5_normalized_RGB_distribution}
\vspace{-15pt}
\end{figure}

 Compared to direct encoding of raw RGB values, encoding residuals using the same size INR results in better object reconstruction quality, as shown in Fig.~\ref{fig:12_more_visualization}. This improvement is attributable to differences in information entropy. The entropy $H$ of a set of random variables $X: \{x_1, x_2, ..., x_n\}$ encoded by a neural network is defined as: $H(X) = -\sum_{i=1}^{n} P(x_i) \cdot \log_2 P(x_i)$.
Smaller $H(x)$ indicates less complex information, facilitating simpler encoding. We compare the distributions of raw and residual RGB values, as illustrated in Fig.~\ref{fig:5_normalized_RGB_distribution}. The normalized raw RGB values display a broader distribution, whereas the normalized residual RGB values cluster around the center value. This concentration implies a higher probability of occurrence near this center value, increasing $P(x_i)$ and thereby reducing $H(x)$. Existing mathematical analyses~\cite{shwartz2017opening, tishby2015deep, achille2019information} suggest that learning targets with lower entropy allow a neural network of the same size to achieve higher learning accuracy.
\begin{figure}[t!]
\centering
\subfigure{\includegraphics[width=1.00\linewidth]{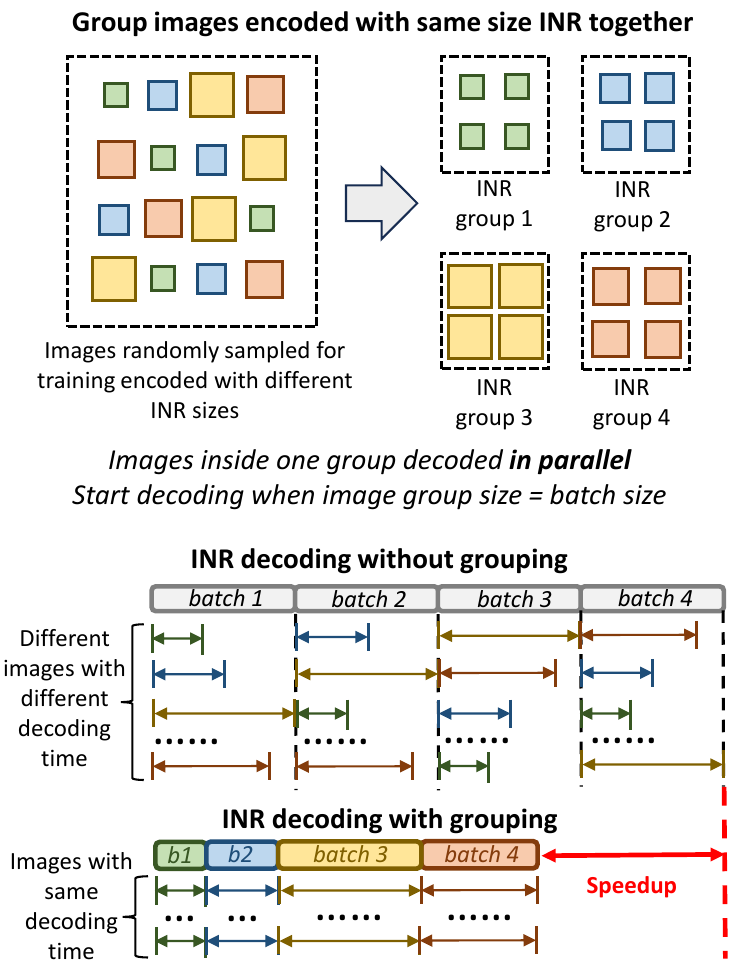}}
\vspace{-20pt}
\caption{Parallelized decoding with balanced workload distribution utilizing INR grouping. }
\label{fig:6_INR_grouping}
\end{figure}

\subsection{Parallelized and balanced decoding}


\subsubsection{CPU-free INR decoding on device} Images compressed to INR format are decoded on edge devices for on-device learning. INR decoding involves neural network inference where Rapid-INR accepts image pixel coordinates and NeRV utilizes video frame temporal indices as inputs. Both of them output image RGB values. Before training starts, all INR weights are transferred once from device storage to device memory in tensor format. This minimizes frequent data exchanges between the device storage and memory during detection backbone training, enabling CPU-free training without the need for a complex software stack. The freed CPU resources can be allocated to other control tasks on the edge device. This hardware efficiency is possible because the dataset, once compressed via INR, is significantly smaller than its JPEG equivalent. INR decoding begins with the background INR, which provides a low-quality RGB reconstruction of the image. Subsequently, the object INR is decoded to retrieve the residual RGB value of the object. By combining this residual with the object RGB value decoded from the background INR, we obtain the final INR reconstructed image, featuring a high-quality object with a low-quality background.


\subsubsection{INR grouping} As shown in Fig.~\ref{fig:6_INR_grouping}, object detection training process involves randomly sampling images from the dataset to form a batch, which helps improve convergence. To facilitate faster decoding, images in INR format within a batch should be decoded in parallel on the device. However, each image is encoded with object INRs of varying sizes, and the background INRs developed upon NeRV also differ in size. These varying sizes result in different decoding latencies. Fig.~\ref{fig:6_INR_grouping} demonstrates that when decoding a batch of images in parallel, the latency required depends on the image with the largest INR, leading to an unbalanced workload on device. To enhance INR decoding speed and thus accelerate on-device training, we propose grouping images with the same-sized INRs together. INR grouping ensures uniform decoding speeds within a batch, balancing the workload and improving the overall training speed.

\begin{figure}[t!]
\centering
\subfigure{\includegraphics[width=1.00\linewidth]{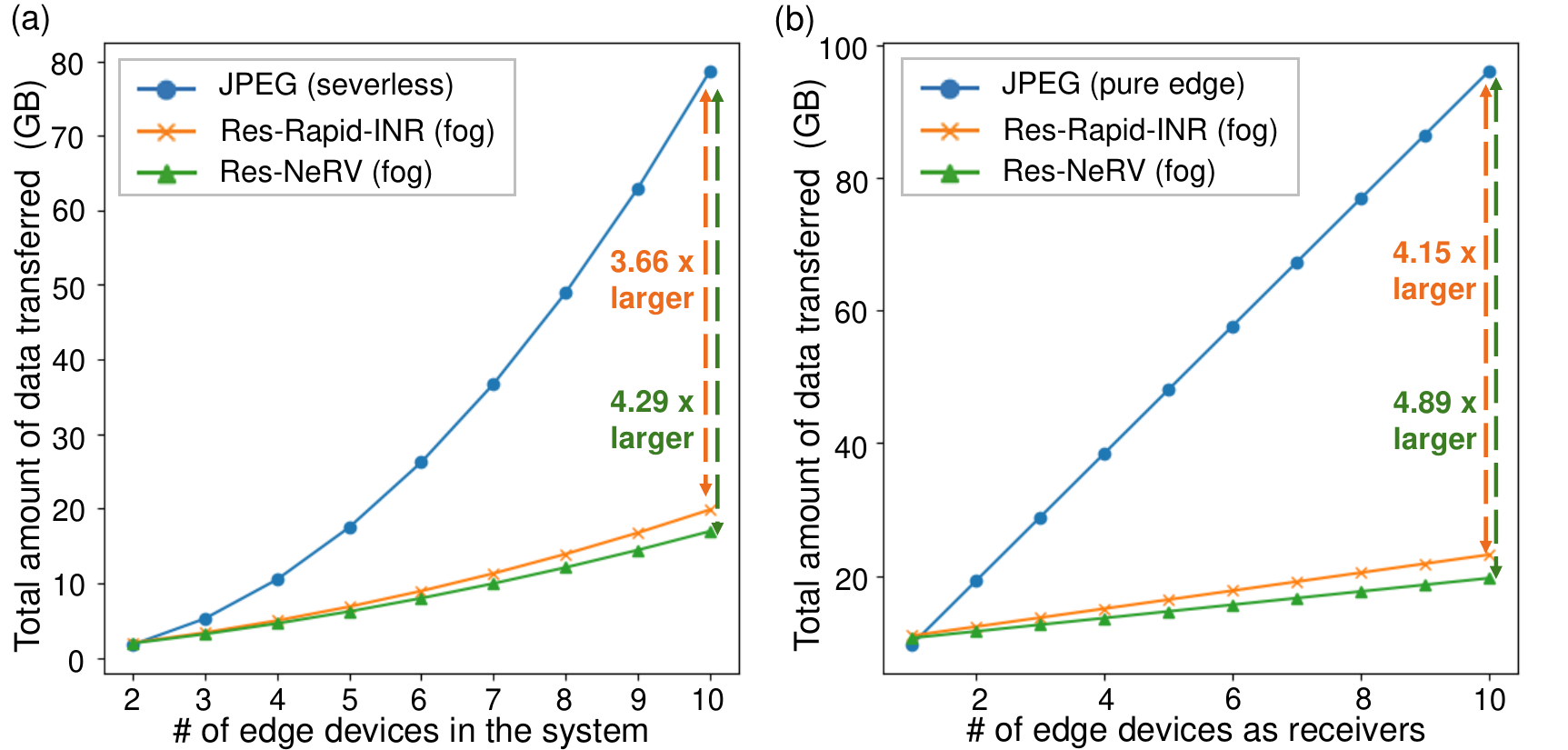}}
\vspace{-20pt}
\caption{(a) Total amount of data transmission within the network with varied number of edge devices, assuming all-to-all communication among them. (b) Total amount of data transmission when each edge device communicates with a varied number of receiver devices, within a network of 11 edge devices.}
\label{fig:7_amount_of_data_transmission}
\end{figure}

\section{Multi-Device Communication Modeling}

We develop a mathematical model to explore the optimal data communication and compression strategies across multiple devices
for fog computing. We aim to minimize communication in the system, i.e., 
transmitting JPEG images to fog nodes for INR compression or direct image transmission to other edge devices in JPEG. Additionally, we investigate whether training at edge or training at fog node is more communication beneficial considering varying numbers of images used for training.

\subsection{Communication modeling}
Our serverless edge computing model is used for data transmission occurring solely among edge devices. Assuming a system comprises $k$ edge devices, with each device transmitting the amount of data $m_i$ to $n_i$ receivers. The total data transmitted within the serverless edge computing system, denoted as $D_s$, is given by: $D_s = \sum_{i=1}^{k} n_i \cdot m_i$.




\begin{figure*}[t!]
\centering
\subfigure{\includegraphics[width=1.00\linewidth]{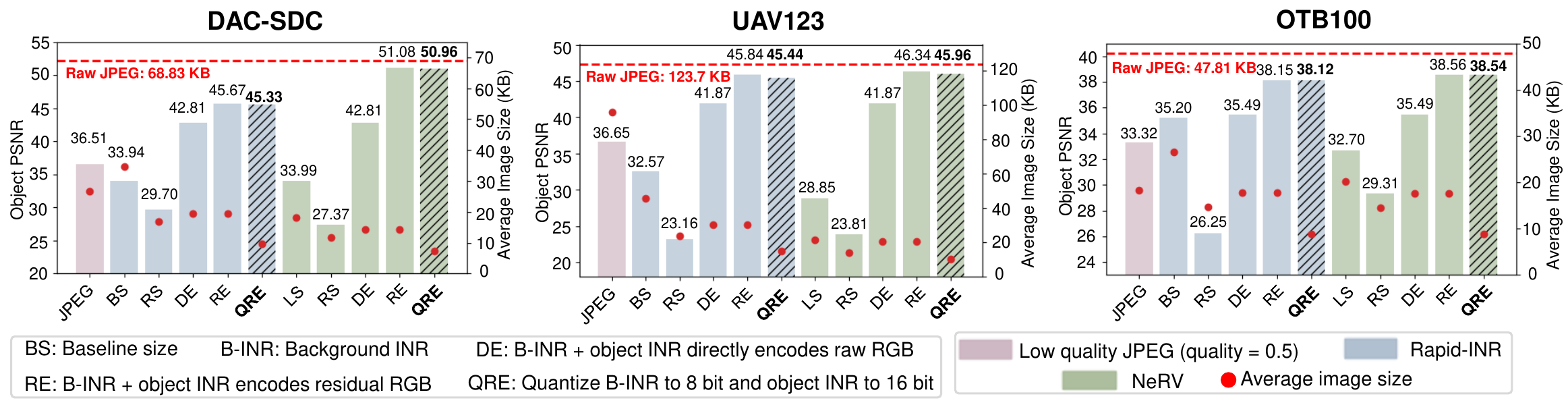}}
\vspace{-20pt}
\caption{Object PSNR relative to the average image size across different compression techniques. Unless otherwise specified, both baseline INR and background INR are quantized to 16 bits. Additionally, we present the average raw image size in JPEG format for three datasets. We choose to quantize the background INR to 8 bits and the object INR to 16 bits, as depicted in the shaded bar graph. }
\label{fig:8_Object_PSNR_INR_Size}
\end{figure*}


Unlike serverless edge computing, fog computing networks also account for data communication between the fog node and edge devices, in addition to interactions among the edge devices themselves. In fog computing, edge devices upload images in JPEG format to the fog node for INR compression. We define the INR compression rate, $\alpha$, as the ratio of the compressed size to the original JPEG size: $\alpha = \frac{\text{INR Size}}{\text{JPEG Size}}$. Assume $k_1$ edge devices choose to upload their images for INR compression and subsequent broadcasting by the fog node, while the remaining devices transmit their images directly in JPEG format to their respective receivers. The total data transmission in this fog computing network, $D_f$, is then calculated as: $D_f = M_1 + M_2 + M_3 = \sum_{i=1}^{k_1} n_i \cdot (\alpha m_i) + \sum_{i=1}^{k_1} m_i + \sum_{i=k_1+1}^{k} n_i \cdot m_i$,
where $M_1$ represents the data broadcast by the fog node after INR compression, $M_2$ is the total amount of data uploaded from edge devices to the fog node, and $M_3$ accounts for the data directly exchanged among edge devices.


\subsection{Optimal communication strategy exploration}


\textbf{INR compression or JPEG?}
Our objective is to minimize the total data transmitted for on-device learning at the edge by leveraging INR compression in fog computing. We aim for $D_f < D_s$ to reduce communication costs. The difference between $D_s$ and $D_f$ is given by: $D_s - D_f = \sum_{i=1}^{k_1} m_i \cdot \left[(1 - \alpha) \cdot n_i - 1\right]$.
For $D_f$ to be minimal, each term in the summation $(1 - \alpha) \cdot n_i - 1$ must be positive. From our mathematical analysis, transferring image data to the fog node for INR compression proves more communication-efficient than directly sending JPEG images to receivers, provided the number of receiving devices $n_i$ for each edge device satisfies: $n_i > \frac{1}{1 - \alpha}$.
This condition ensures that the benefits of INR compression outweigh the extra costs of uploading JPEG images to the fog node. Fig.~\ref{fig:7_amount_of_data_transmission} illustrates the improved communication efficiency achieved by using fog computing with INR compression, in comparison to serverless edge computing.

\begin{figure*}[t!]
\centering
\subfigure{\includegraphics[width=1.00\linewidth]{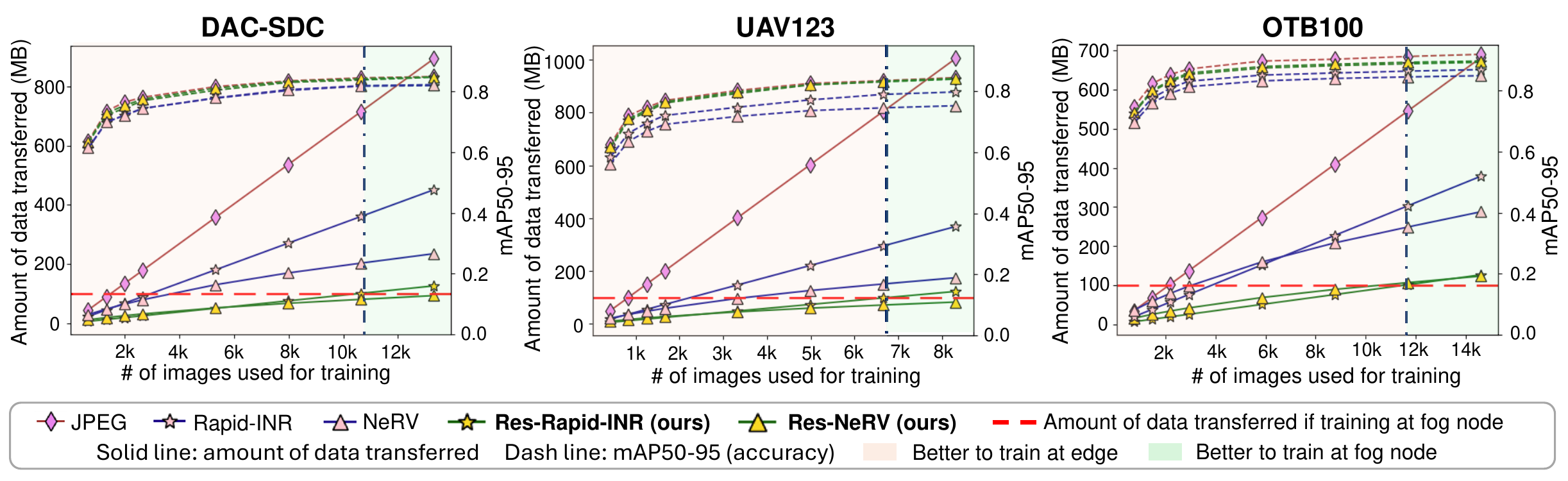}}
\vspace{-20pt}
\caption{The relationship between training accuracy, amount of data transferred between the fog node and edge device, and the number of images used for training across various compression techniques. We identify the most communication-efficient training strategy for different image quantities. Training at fog node transfers YOLOv8 model weights.}
\label{fig:9_online_learning_accuracy}
\end{figure*}

\textbf{Training at fog node or at edge?} 
The enhanced computational capabilities of the fog node allow for the possibility of transferring model weights to the fog node for training and subsequently transferring the trained weights back to the edge devices. The decision between transferring model weights or images depends on which is larger: the amount of data required for on-device learning or the twice of model size. Typically, in on-device learning scenarios, the amount of new data used for fine-tuning is smaller than twice of the model size, leading to transferring INR-formatted images to the edge for training is more communication beneficial. However, in cases where the new task significantly deviates from the original task used for model training, it is more communication efficient to transfer the model weights back to the fog node for retraining.


\section{Experiment Results}

\subsection{Experiment settings}

\subsubsection{Datasets and platforms for evaluation} 
To demonstrate the efficiency of Residual-INR in fog computing for on-device learning, we focus on single object detection, employing YOLOv8~\cite{ultralytics2023} middle size (98.8 MB) as the detection backbone. We evaluate Residual-INR using three datasets: DAC-SDC~\cite{xu2019dac}, UAV123~\cite{mueller2016benchmark}, and OTB100~\cite{wu2013online}. These three datasets consist of multiple video sequences, each representing a specific object category and consisting of continuous video frames stored in JPEG format. This setup allows for two INR compression options: treating video frames as independent images using Rapid-INR, or encoding them as continuous video sequences with NeRV. The image decoding and training latency measurement are conducted on real hardware systems, using Intel 6226R CPU and NVIDIA A6000 GPU to simulate the edge computing device. To demonstrate the training speedup offered by Residual-INR, we compare our INR training pipeline with two prevalent training frameworks: PyTorch~\cite{PyTorch}, which utilizes CPU for data loading and JPEG decoding, and DALI~\cite{DALI}, which employs a mixed CPU and GPU setup for accelerated JPEG decoding. To show the system-level benefits of Residual-INR in fog computing for reduced communication, our analysis relies on simulation results derived from our mathematical model. We set the wireless communication bandwidth as 2MB/s in our experiments.

\subsubsection{Detection backbone training and INR encoding} 
To simulate on-device learning, we initially randomly selected half of the video sequences from each dataset to train YOLOv8 and develop a pretrained model. Subsequently, we select new video sequences from the remaining half to fine-tune this pretrained model, evaluating detection accuracy on those new videos. The fine-tuning takes 10 epochs, which is sufficient for the detection backbone to converge with the new data. We adhere to the default settings of YOLOv8 for other training hyper-parameters. Notably, INR grouping is employed when training with images decoded by Residual-INR, but not with images decoded by Rapid-INR or the NeRV baseline. We use varying sizes of INR for encoding. The detailed architectures of the Rapid-INR series are presented in Tab.~\ref{tab:Rapid-INR arch}, and those of the NeRV series are shown in Tab.~\ref{tab:NeRV arch}.

\begin{table}[t!]
\footnotesize
\centering
\caption{Res-Rapid-INR (background INR + object INR), and Rapid-INR baseline configuration details (layer count 
$\times$ hidden dimension) of MLP. }
\vspace{-10pt}
\begin{tabular}{|c|c|c|c|}
\hline
\textbf{} & \textbf{DAC-SDC} & \textbf{UAV123} & \textbf{OTB100} \\ \hline
\textbf{Background INR} & 10$\times$30 & 10$\times$36 & 10$\times$28 \\ \hline
\textbf{Object INR} & \makecell{3$\times$10, 3$\times$15, \\ 5$\times$17, 5$\times$24} & \makecell{3$\times$15, 5$\times$17, \\ 5$\times$24, 6$\times$28} & \makecell{3$\times$15, 5$\times$17, \\ 5$\times$24, 6$\times$28} \\ \hline
\textbf{\makecell{Rapid-INR}} & 16$\times$48 & 16$\times$55 & 14$\times$45 \\ \hline
\end{tabular}
\label{tab:Rapid-INR arch}
\vspace{-15pt}
\end{table}

\subsection{Object reconstruction quality}
We evaluate object reconstruction quality across various encoding methods, including different JPEG qualities and INR configurations, using PSNR as the metric for object region quality. Additionally, we analyze the average image sizes produced by these methods. Fig.~\ref{fig:8_Object_PSNR_INR_Size} indicates that after quantizing the background INR to 8 bits and the object INR to 16 bits, both Res-Rapid-INR and Res-NeRV significantly outperform the Rapid-INR and NeRV baselines, as well as low-quality JPEG, in terms of PSNR of object. With image sizes ranging from 8.3\% to 18.4\% of the original JPEG, Residual-INR achieves a PSNR over 38, closely approximating the quality of the raw RGB. Furthermore, for the same average image size, residual encoding provides superior object quality compared to direct RGB encoding.

\begin{table}[t!]
\footnotesize
\centering
\caption{NeRV background INR (B-S: small, B-M: medium and B-L: large), object INR (O-INR) and NeRV baseline (S: small, M: medium, L: large) configuration details. M represents two hidden layer dimensions of MLP (dim 1, dim 2) of MLP; C represents the number of channels for the first two layers and the middle two layers of the CNN (channel1, channel2). The size of object INR MLP is represented by (layer count $\times$ hidden dimension).}
\vspace{-10pt}
\begin{tabular}{|c|c|c|c|}
\hline
\textbf{} & \textbf{DAC-SDC} & \textbf{UAV123} & \textbf{OTB100} \\ \hline
\textbf{B-S} & \makecell{M: (512, 3744) \\ C: (26, 96)} & \makecell{M: (512, 3744) \\ C: (26, 96)} & \makecell{M: (256, 2304) \\ C: (16, 96)} \\ \hline
\textbf{B-M} & \makecell{M: (512, 8352) \\ C: (58, 96)} & \makecell{M: (512, 8352) \\ C: (58, 96)} & \makecell{M: (512, 3744) \\ C: (26, 96)} \\ \hline
\textbf{B-L} & \makecell{M: (512, 16128) \\ C: (112, 96)} & \makecell{M: (512, 16128) \\ C: (112, 96)} & \makecell{M: (512, 8352) \\ C: (58, 96)} \\ \hline
\textbf{O-INR} & \makecell{3$\times$10, 3$\times$15, \\ 5$\times$17, 5$\times$24} & \makecell{3$\times$15, 5$\times$17, \\ 5$\times$24, 6$\times$28} & \makecell{3$\times$15, 5$\times$17, \\ 5$\times$24, 6$\times$28} \\ \hline
\textbf{NeRV-S} & \makecell{M: (768, 8352) \\ C: (58, 192)} & \makecell{M: (768, 8352) \\ C: (58, 192)} & \makecell{M: (768, 3744) \\ C: (26, 96)} \\ \hline
\textbf{NeRV-M} & \makecell{M: (768, 16128) \\ C: (112, 192)} & \makecell{M: (768, 16128) \\ C: (112, 192)} & \makecell{M: (768, 8352) \\ C: (58, 96)} \\ \hline
\textbf{NeRV-L} & \makecell{M: (768, 28224) \\ C: (196, 192)} & \makecell{M: (768, 28224) \\ C: (196, 192)} & \makecell{M: (768, 16128) \\ C: (112, 96)}\\ \hline
\end{tabular}
\label{tab:NeRV arch}
\vspace{-15pt}
\end{table}


\begin{figure}[t!]
\centering
\vspace{-10pt}
\subfigure{\includegraphics[width=1.00\linewidth]{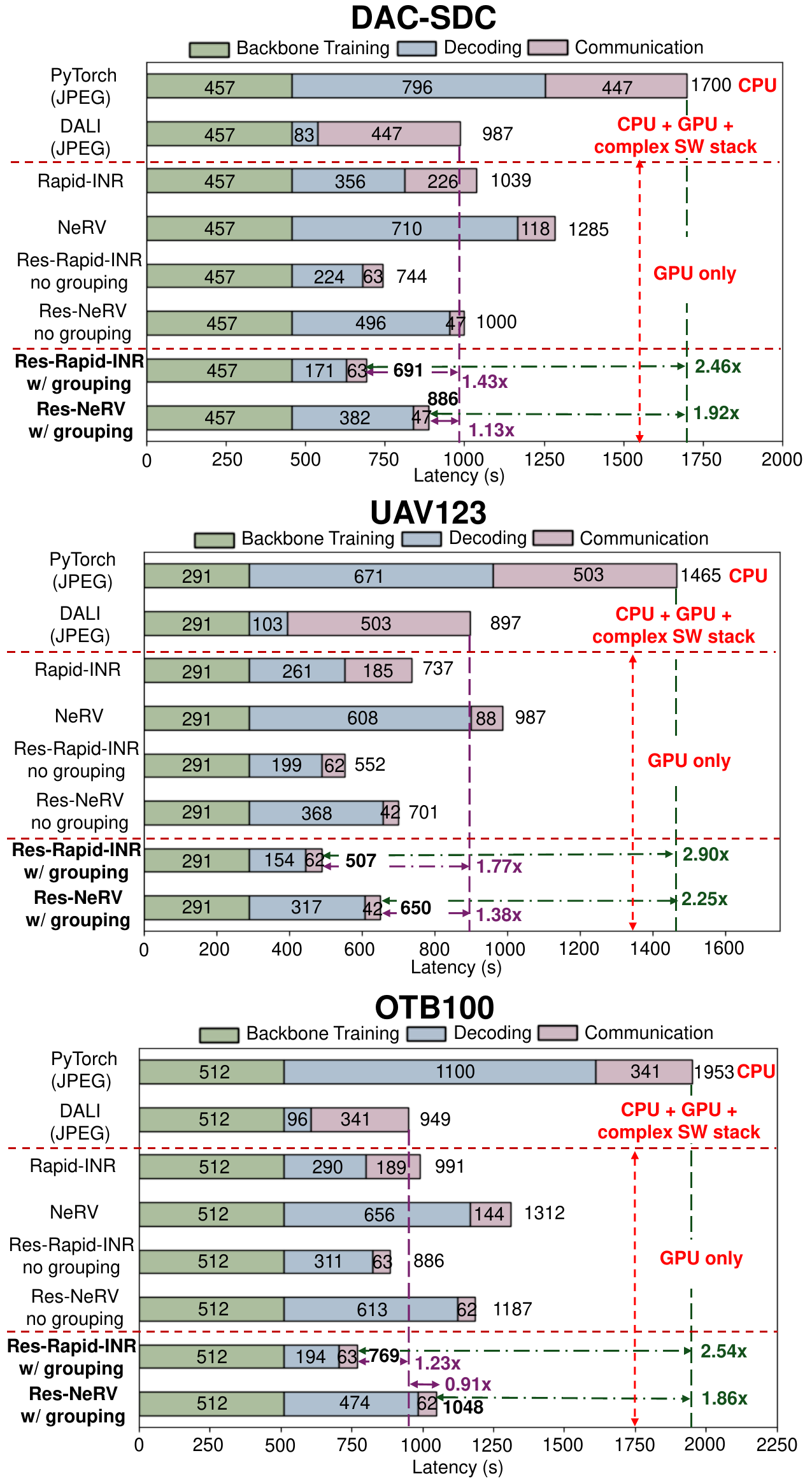}}
\vspace{-25pt}
\caption{Detailed latency breakdown for training at the edge in fog computing network, with our selected approach highlighted in bold. }
\label{fig:10_hw_speedup}
\vspace{-18pt}
\end{figure}

\subsection{Detection backbone training accuracy with amount of data transmission}


We fine-tune the YOLOv8 model using varying numbers of images sampled from new selected video sequences. As shown in Fig.~\ref{fig:9_online_learning_accuracy}, both the training accuracy (mAP50-95) and the data volume transferred from the fog node to a single edge device increase with the number of images used for training. Res-Rapid-INR and Res-NeRV significantly reduce the amount of data transferred compared to JPEG, Rapid-INR, and NeRV baselines, while achieving training accuracy comparable to that of raw JPEG, and higher than Rapid-INR and NeRV. This demonstrates that Residual-INR can significantly alleviate communication traffic without compromising training accuracy.

Additionally, we compare the amount of data transmission between training at edge and training at fog node. The data transferred is double the model size, as the model is retrieved from and then sent back to the edge device after training. Assuming the YOLOv8 model is quantized to 16 bits for transferring, Fig.~\ref{fig:9_online_learning_accuracy} illustrates that using Res-Rapid-INR in the pink region results in lower data transmission amount for training at the edge. In the green region, it is more beneficial to transfer YOLOv8 weights back to fog node for training.

\subsection{Training speedup and detailed breakdown}
We conduct an ablation study to show the power of Residual-INR in reducing wireless data transmission time and the advantages of INR grouping in minimizing INR decoding time. The detailed breakdown of training time, including detection backbone training, image decoding, and transmission time, is depicted in Fig.~\ref{fig:10_hw_speedup}. Because of reduced data transmission and accelerated decoding via INR grouping, the end-to-end training time for Res-Rapid-INR and Res-NeRV showed a speedup of up to 2.9 $\times$ and 2.25 $\times$, respectively, compared to a PyTorch training pipeline with JPEG images decoded on a single-thread CPU. This speedup reaches 1.77 $\times$ and 1.38 $\times$, respectively, compared to a DALI training pipeline with GPU-accelerated JPEG decoding. INR grouping yields an average speedup of 1.40 $\times$ for Res-Rapid-INR and 1.25 $\times$ for Res-NeRV across three datasets.


\begin{figure}[t!]
\centering
\vspace{-0pt}
\subfigure{\includegraphics[width=1.00\linewidth]{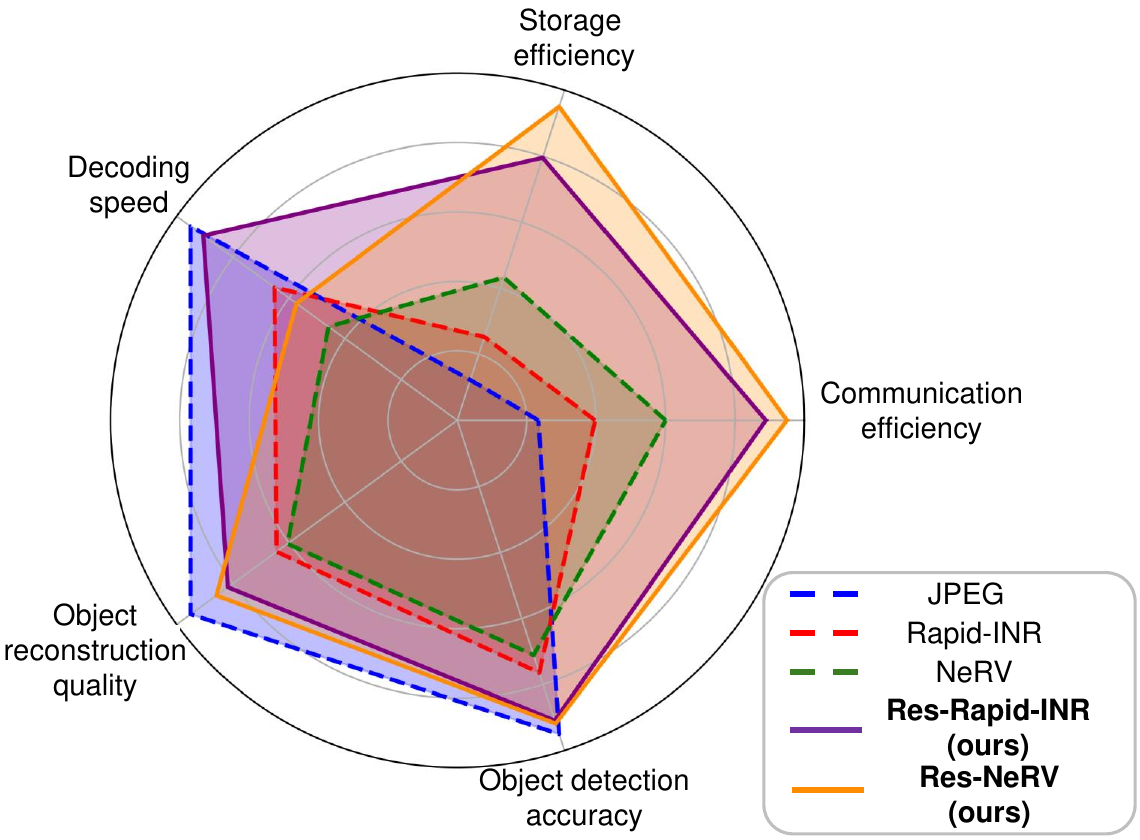}}
\vspace{-20pt}
\caption{Radar graph comparing the advantages and disadvantages of various compression techniques.}
\label{fig:11_radar_graph}
\vspace{-18pt}
\end{figure}

\subsection{Summary of different compression techniques}
To do a comprehensive evaluation on various compression techniques, we compare JPEG, Rapid-INR, NeRV, Res-Rapid-INR, and Res-NeRV across multiple metrics, as illustrated in a radar graph (Fig.~\ref{fig:11_radar_graph}). While JPEG offers the highest object quality and detection accuracy, it incurs significant storage and communication costs. Additionally, JPEG decoding on CPUs is slow, and GPU-accelerated decoding requires a complex software stack.
In contrast, Residual-INR significantly enhances storage and communication efficiency with minimal influence on object quality and detection training accuracy compared to JPEG. Additionally, it provides faster decoding speeds and higher detection accuracy than the Rapid-INR and NeRV baselines.



\section{Conclusion}

In this paper, we propose \textbf{Residual-INR}, a communication efficient fog on-device learning framework that utilizes region importance aware INRs for image and video compression. By separately encoding the object and background of an image or video frame with a smaller INR for low-quality background and a tiny INR for high-quality object encoding, Residual-INR compresses images to sizes 5.4 to 12.1 $\times$ smaller than JPEG. Moreover, Residual-INR reduces data transmission volumes by 3.43 to 5.16 $\times$ across a fog computing network of 10 edge devices compared to serverless edge computing. Furthermore, Residual-INR enables CPU-free on device training without the need for a complex software stack. It can achieve up to a 2.9 $\times$ speedup compared with a PyTorch training pipeline using JPEG and up to 1.77 $\times$ faster than accelerated training pipeline with JPEG using DALI.

\section{Acknowledgements}
This work is partially supported by Samsung, Cisco, and the National Science Foundation under Grant ENG-ECCS-2202310.



\bibliography{reference}

\end{document}